\documentclass{article} 

\usepackage{graphicx} 
\usepackage{amsmath,amsfonts,amssymb,amsthm,epsfig,epstopdf,url,array}
\usepackage{booktabs} 
\usepackage{url}    
 
\usepackage[utf8]{inputenc} 
\usepackage[T1]{fontenc}    
\usepackage{url}            
\usepackage{booktabs}       
\usepackage{amsfonts}       
\usepackage{nicefrac}       
\usepackage{microtype}      
\usepackage{xcolor}         
\usepackage{graphicx}  
\usepackage[font=small]{caption} 
\usepackage{subcaption}
\usepackage{makecell}
\usepackage{framed}  
\usepackage{amsmath} 
\usepackage{amssymb}
\usepackage{bm}     
\usepackage{algorithm}
\usepackage{algorithmic}
\usepackage{color}
\usepackage{xparse}
\usepackage{mathtools}     
\usepackage{wasysym}
\usepackage[flushleft]{threeparttable}
\usepackage{multirow}
\usepackage{enumitem}

\usepackage{hyperref}

\usepackage[accepted]{icml2023}

\usepackage[capitalize,noabbrev]{cleveref}

\theoremstyle{plain}

\theoremstyle{definition}

\theoremstyle{remark}

\usepackage[textsize=tiny]{todonotes}

\icmltitlerunning{Test-Time Style Shifting: Handling Arbitrary Styles in Domain Generalization}

\begin{document}

\twocolumn[
\icmltitle{Test-Time Style Shifting: Handling Arbitrary Styles in Domain Generalization}



\icmlsetsymbol{equal}{*}

\begin{icmlauthorlist}
\icmlauthor{Jungwuk Park}{equal,kaist}
\icmlauthor{Dong-Jun Han}{equal,purdue}
\icmlauthor{Soyeong Kim}{kaist}
\icmlauthor{Jaekyun Moon}{kaist}
\end{icmlauthorlist}

\icmlaffiliation{kaist}{School of Electrical Engineering, Korea Advanced Institute of Science and Technology (KAIST), South Korea.}
\icmlaffiliation{purdue}{Purdue University, USA}

\icmlcorrespondingauthor{Jaekyun Moon}{jmoon@kaist.edu}

\icmlkeywords{Machine Learning, ICML}

\vskip 0.3in
]



\printAffiliationsAndNotice{\icmlEqualContribution} 

\begin{abstract}
In domain generalization (DG), the target domain is
unknown when the model is being trained, and
the trained model should successfully work on
an arbitrary (and possibly unseen) target domain
during inference. This is a difficult problem, and despite active studies in recent years, it remains a great challenge.
In this paper, we take a simple yet effective approach to tackle this issue. We propose \textit{test-time style shifting}, which shifts the style of  the test sample (that has a large style gap with the source domains) to the nearest    source domain that the model is already familiar with, before making the prediction. This strategy enables the model to handle any target domains with arbitrary style statistics, without additional model update at test-time.  Additionally, we propose  \textit{style balancing}, which provides a great platform for maximizing the advantage of  test-time style shifting by handling the DG-specific imbalance issues.   The proposed ideas are easy to implement and   successfully work in conjunction with various  other DG schemes.  Experimental results on different datasets
show the effectiveness of our methods.


\end{abstract}

\section{Introduction}
The huge success of deep convolutional neural networks (CNNs) relies on the assumption that the \textit{domains} of the training data and the test data are the same. However, this assumption does not hold in practice. For example, in self-driving cars, although we may only have train images on sunny days and
foggy days during training (source domains), we would have to make predictions for images on snowy days
during testing (unseen target domain). 
Due to the practical significance of this problem setup,  domain generalization (DG) is receiving considerable attention nowadays.

Given a training set that consists of  multiple (or a single) source domains, the goal of DG 
is to achieve generalization capability to predict well on an  arbitrary target domain. Existing works  tackle this   problem via meta-learning \cite{li2019episodic,    li2018learning, zhao2021learning},  data augmentation \cite{nam2021reducing, shankar2018generalizing,   yue2019domain, zhou2020learning} or domain alignment \cite{li2018domain, li2018deep, li2018domain, erfani2016robust}.  
Recently, motivated by the observations  \cite{huang2017arbitrary, li2021feature, dumoulin2017learned} that the domain characteristic of  data has a strong correlation with the   feature statistics (or style statistics) of the early layers of CNNs, 
the authors of \cite{zhou2021domain, li2022uncertainty, zhang2022exact, kang2022style} proposed to generate new  style statistics  during training via style augmentation.  

\begin{figure}
  \centering
  \includegraphics[width=0.47\textwidth]{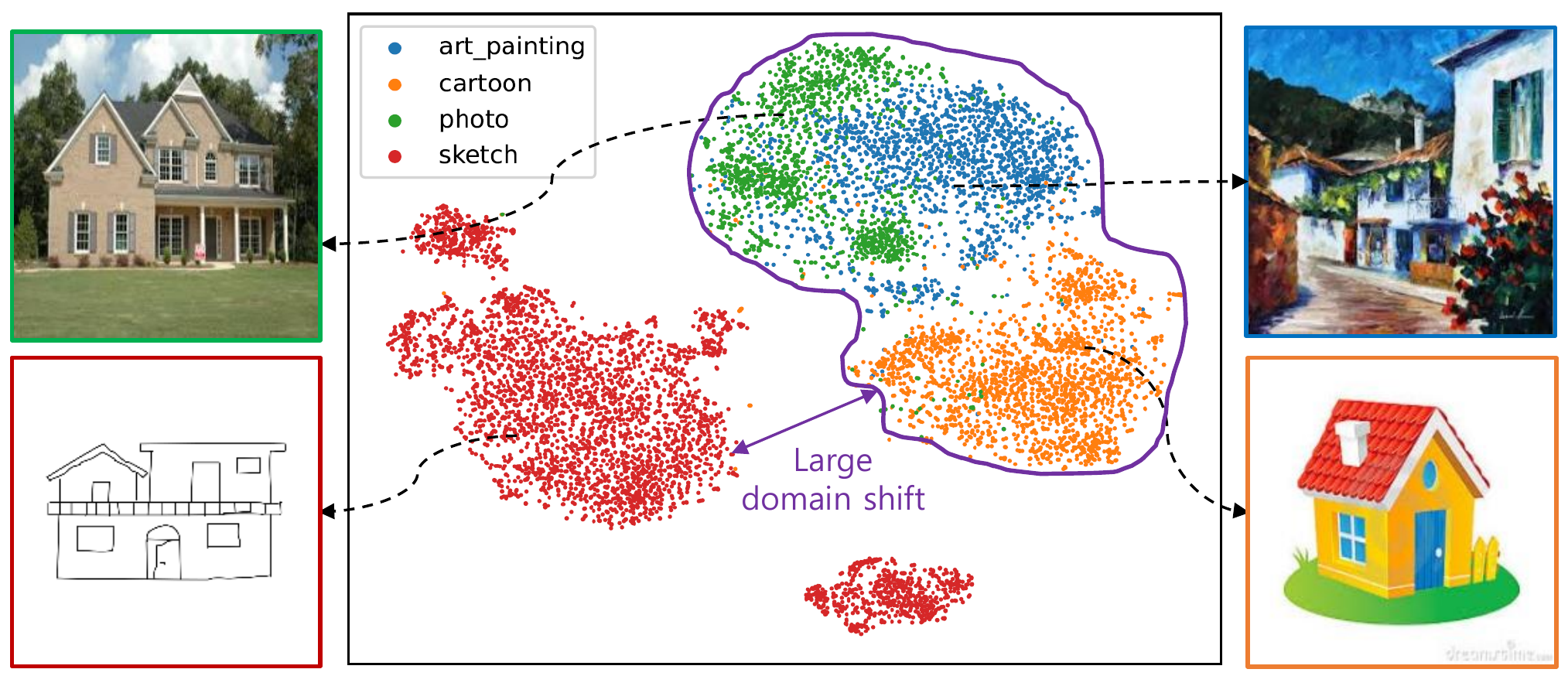}   
  \caption{t-SNE  of concatenated  feature-level style statistics $\Phi =[\mu, \sigma]$ of  samples, obtained from the output of second residual block of ResNet-18. Samples are clustered based on domain characteristics. Sketch domain has a large style gap with other domains, resulting in low   model accuracy when Sketch is the target domain. Our test-time style shifting tackles this scenario by shifting the style  of the test sample  to the nearest source domain that the model is already well-trained on.}  
  \label{fig:tsne}
\end{figure}

However, DG is still regarded   as a challenging problem  since the target domain is unknown during training, and the trained model should be able to handle \textit{arbitrary}, and \textit{possibly unseen} target domains during inference; the target domain could have a significantly large discrepancy with the source domains due to domain shift, limiting the prediction performance. As an example, consider the well-known PACS dataset \cite{li2017deeper}  in Fig. \ref{fig:tsne}, which shows the t-SNE of feature-level styles statistics of the samples. It can be seen that the Sketch domain has a large style gap with other source domains, which results in limited performance when Sketch  domain becomes the target.

\textbf{Contributions.} In this paper, we take a simple yet effective approach to improve  the DG performance when there is a significant domain shift between the source  and   target domains. Specifically,  in order to handle arbitrary target domains during inference, we propose \textit{test-time style shifting}, which shifts  the style of the test sample (that has a   large style gap  with the source domains) to  the nearest  source domain  that the model is already familiar with, before making the prediction. Note that our scheme only performs style shifting in the style-space and thus does not require any model updates at test-time. Moreover,  our test-time style shifting does not require additional changes in the model architecture or the objective function, making our scheme to be more compatible with any tasks/models.

In order to maximize the effectiveness of test-time style shifting, the model should be well-trained   on the styles of the source domains. Motivated by this, we also propose   \textit{style balancing}, which provides a great platform to increase the potential of test-time style shifting by handling the DG-specific imbalance issues. Note that in DG scenarios with multiple domains,  the  imbalance issues have different characteristics compared to the traditional class imbalance problem in   a single domain; when a specific domain lacks certain classes,   it turns out, as will seen    in Section  \ref{sec:further},  that  
existing methods based on resampling or reweighting  fail to handle these DG-specific imbalance issues. Our proposed style balancing handles this issue by choosing the sample that has similar style statistics to other samples (and thus has a similar role compared to others) in the same domain, and converting the style of this sample to another domain; this improves the domain diversity per classes during training by compensating for the missing classes in each domain. 

Our  test-time style shifting  and style balancing work in a highly complementary fashion; style balancing plays a key role in improving the performance of test-time style shifting by exposing the model to various styles per classes during training. Moreover,  removing one of these components   can   degrade  the performance in practice having (i) DG-specific imbalance issues and (ii) large domain shift between source and target at the same time.  Our solution is compatible  with  not only the style-augmentation based DG schemes (e.g., MixStyle, DSU, EFDMix) that   operate in the style space as ours, but also other DG ideas relying on domain alignment or meta-learning.  Extensive experimental results on various DG benchmarks show the improved performance of our scheme over existing  methods.


\section{Related Works}  

\textbf{DG with style augmentation.} DG has been actively studied for the past few years using meta-learning \cite{li2019episodic, chen2022compound, du2020learning, li2018learning, zhao2021learning},  data augmentation \cite{nam2021reducing, shankar2018generalizing,   yue2019domain, zhou2020learning},  domain alignment \cite{li2018domain, li2018deep, li2018domain, erfani2016robust} and so on. 
Recently, various   style  augmentation methods such as MixStyle \cite{zhou2021domain}, DSU  \cite{li2022uncertainty}, Style Neophile \cite{kang2022style} and EFDMix \cite{zhang2022exact} have been also proposed. As in our solution, style augmentation based DG schemes can be simply applied to any tasks/models and operate in the style space defined with style statistics. However, although these DG approaches  explore new styles via style-augmentation, these methods are not able to cover an arbitrary style  that has a large style gap  with the source domain. Also, the performance of these methods could be potentially limited in practice with DG-specific imbalance issues; even when combined with existing class-imbalanced solutions, the issue on  the missing classes in each domain cannot be directly handled,     as shown in Section  \ref{sec:experiment}. Our   test-time shifting and style balancing can successfully work in conjunction with  recent style augmentation   strategies (and also with other DG methods) to handle these fundamental issues.




 \textbf{Class-imbalanced learning.} Targeting   class-imbalanced datasets, various    over/down-sampling  strategies  \cite{he2008adasyn, pouyanfar2018dynamic} and loss function modification (e.g., reweighting)  methods  \cite{huang2016learning, shu2019meta, cui2019class} have been proposed.  While these works focus on   class imbalance   within a single domain, in a DG setup with multiple domains, the imbalance issues make the problem more challenging. Especially when a specific domain does not have data samples of certain classes,  these missing classes   cannot be compensated via over/under-sampling or  loss modification strategies.   A recent work \cite{yang2022multi} focused on  a similar multi-domain setup with imbalanced datasets, by defining a new loss function using the distance between representations.  However,  the   loss function proposed in \cite{yang2022multi} does not capture the  classes missing in each domain. 
 Our style balancing module handles this issue by shifting the style statistics of the sample  to another domain, compensating for  the missing classes in each domain.  
  

\textbf{Test-time adaptation.}  Several test-time adaptation methods \cite{wang2020tent, iwasawa2021test, pandey2021generalization, sun2020test, xiao2022learning, zhao2022test} have been recently proposed,   where \cite{pandey2021generalization, iwasawa2021test,  xiao2022learning, zhao2022test}  specifically focused  on DG. In \cite{wang2020tent, iwasawa2021test,  sun2020test}, the authors proposed schemes to update   model parameters during testing. Compared to these works, our test-time style shifting does not require further model update at test time; we simply utilize adaptive instance-normalization (AdaIN)  \cite{huang2017arbitrary} to shift the style of the test sample to the familiar source domain.  
Recently in  \cite{xiao2022learning}, the authors proposed a  method that does not require fine-tuning on target samples at test-time.  
However, this work requires additional networks and perform Monte Carlo sampling for  variational inference, which increases  training costs.  Notably, \cite{pandey2021generalization} proposed to construct a source manifold and projects the feature of the test samples to this source manifold.  Orthogonal to this work focusing on the output of the feature extractor where the data are clustered according to classes (regardless of the domains), we deal with shifting the style statistics at earlier
 layers where the data are clustered according to the domains (regardless of the classes).  Moreover,  our test-time style shifting does not require additional changes in the model architecture or the objective function, making our scheme to be more compatible with any task/models. To the best of our knowledge, our approach is the first work that shifts the   feature-level \textit{style statistics} of the target sample  in the style space at testing.  

 We stress that our style balancing and test-time style shifting  are orthogonal to the aforementioned works in that   we only  shift the style statistics in the style-space during training/testing. Previous works on DG, class-imbalanced learning and test-time adaptation   can work in conjunction with our scheme to  improve the prediction performance further.





\section{Problem Setup}
\subsection{Backgrounds:  Style Augmentation in DG}\label{subsec:back}
Let $x\in \mathbb{R}^{B\times C \times H \times W}$ be a mini-batch of features at a specific layer, where $B$, $C$, $H$, $W$ are the dimensions of mini-batch, channel, height, width, respectively. We also let $\mu(x)\in \mathbb{R}^{B\times C}$ and $\sigma(x)\in \mathbb{R}^{B\times C}$ be the channel-wise mean and standard deviation of each instance within the batch as:
\begin{equation}\label{eq:mu1}
\mu(x)_{b,c} = \frac{1}{HW}\sum_{h=1}^H\sum_{w=1}^Wx_{b,c,h,w},
\end{equation}
\begin{equation}\label{eq:mu2}
\sigma^2(x)_{b,c} = \frac{1}{HW}\sum_{h=1}^H\sum_{w=1}^W(x_{b,c,h,w} -\mu_{b,c}(x))^2. 
\end{equation}
The values $\mu(x)$  and $\sigma(x)$ denote instance-level feature statistics  of $x$. These values   also denote  \textbf{style statistics} since the instance-level feature statistics carry out style information in CNNs \cite{huang2017arbitrary}. Now define new style  statistics $\mu(y)$ and $\sigma(y)$ computed by feature $y$, corresponding to another batch of images.    According to  AdaIN \cite{huang2017arbitrary},  one can generate new features having content $x$ and style $y$ as follows:
\begin{equation}\label{eq:adain}
\text{AdaIN}(x, y) = \sigma(y)\frac{x-\mu(x)}{\sigma(x)} + \mu(y). 
\end{equation} 
Based on AdaIN,  MixStyle \cite{zhou2021domain}  and DSU   \cite{li2022uncertainty} focus on constructing new style  statistics   as $\gamma\frac{x-\mu(x)}{\sigma(x)} + \beta$ to improve generalization, where  $\beta$ and $\gamma$ are the coefficients that determine the  style of the image as in (\ref{eq:adain}). 
MixStyle specifically mixes the style statistics  as $\beta = \lambda \mu(x) + (1-\lambda)\mu(y)$, $\gamma= \lambda \sigma(x) + (1-\lambda)\sigma(y)$ for  $0<\lambda<1$.
 On the other hand, DSU generates new styles by sampling $\beta_{\text{mix}}$ and  $\gamma_{\text{mix}}$  from  Gaussian distributions. 

The authors of  \cite{zhang2022exact}  proposed EFDM to replace AdaIN in (\ref{eq:adain}). By redefining $x\in \mathbb{R}^{HW}$  on a specific sample and a channel, the elements of vector $x$ are reordered in an ascending order as $[x_{\tau_1},x_{\tau_2},\dots,x_{\tau_{HW}}]$, where $x_{\tau_i}\leq x_{\tau_j}$ holds for $i<j$ and $\{x_{\tau_i}\}_{i=1}^{HW}$ are the elements of vector $x$.  The elements of $y$ are similarly reordered as  $[y_{\kappa_1},y_{\kappa_2},\dots,y_{\kappa_{HW}}]$. Then, arbitrary style transfer can be performed as       $\text{EFDM}(x,y)_{\tau_i} = y_{\kappa_i}$ to replace (\ref{eq:adain}), where $\text{EFDM}(x)_{\tau_i}$ is the $\tau_i$-th element of the output.  Based on EFDM, the authors of  \cite{zhang2022exact} also  propose EFDMix, which replaces the concept of AdaIN  in  MixStyle with  EFDM,  in a channel-wise manner as follows: $\text{EFDMix}(x)_{\tau_i} =  \lambda x_{\tau_i} + (1-\lambda)y_{\kappa_i}$.  


 \begin{figure*}[t]
\centering
    \centerline{\includegraphics[width=150mm]{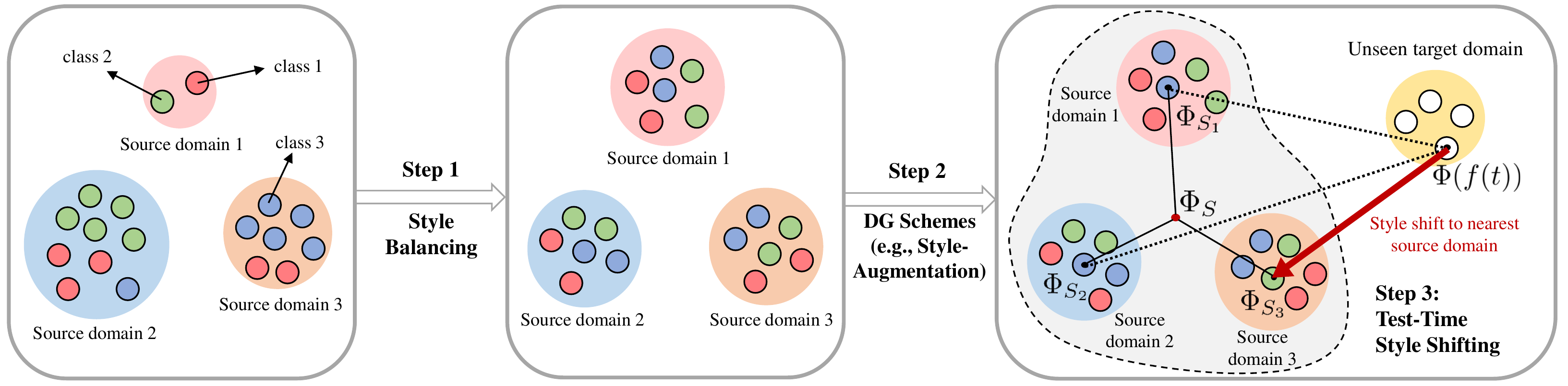}}   
  \caption{\small  An overview of proposed idea. Given imbalanced feature statistics at a specific layer, style balancing is first performed  to balance the style statistics across all domains. Then, a specific DG scheme  (e.g., style-augmentation) can be adopted for training. At testing, the style of the  test sample (far from the source domains) is shifted  to the nearest source domain  based on our test-time style shifting. The style balancing module and the test-time style shifting module can be flexibly applied at any layers of the backbone network. } 
  \label{fig:algorithm}
\end{figure*}

\subsection{Problem Formulation}
\textbf{Notations.} Let $N$ be the number of source domains and $S_{n}$ be the set of train samples in source domain $n$, where $S= \cup_{n=1}^NS_n$ is the overall train set. Let $S_{n,k}$ be the set of train samples in   domain $n$ labeled as class $k$ satisfying $S_n = \cup_{k=1}^KS_{n,k}$, where $K$ is the number of classes. Given a  sample $s\in S$, let $f(s)\in \mathbb{R}^{C \times H \times W}$ be the encoded features at a specific layer. We define $\mu(f(s))\in \mathbb{R}^C$ and $\sigma(f(s))\in \mathbb{R}^C$ as the channel-wise mean and standard deviation of $f(s)$, similar to   (\ref{eq:mu1}) and  (\ref{eq:mu2}).  Related to the notations in Section \ref{subsec:back}, we have $x=[f(s_1),f(s_2),\dots , f(s_B)]$, $\mu(x)=[\mu(f(s_1)),\mu(f(s_2)),\dots , \mu(f(s_B))]$, $\sigma(x)=[\sigma(f(s_1)),\sigma(f(s_2)),\dots , \sigma(f(s_B))]$ where $B$ is the  batch size. For any set $A\subseteq S$, we also define the mean of style statistics in set $A$ as 
$ \mu_A = \frac{1}{|A|}\sum_{s\in A}\mu(f(s))$, $\sigma_A = \frac{1}{|A|}\sum_{s\in A}\sigma(f(s))$.  
Given a set $A$ and corresponding $\mu_A$, $\sigma_A$, the concatenation of these two are defined as
\begin{equation}\label{eq:defn}
\Phi_A = [\mu_A, \sigma_A].
\end{equation}
Similarly, we define $\Phi(f(s)) = [\mu(f(s)), \sigma(f(s))]$ for any sample $s$. Using these notations,    we can formally state the   issue and goal  of the paper as follows.

\textbf{Issue and goal.} Let $t$ be the test sample in the  target domain.  In practice, the   style gap between source and target domains could be large, i.e.,  $\|\Phi_{S_n} - \Phi(f(t))\|$ is large for all $n\in\{1,2,\dots,N\}$ (see Sketch domain in Fig. \ref{fig:tsne}). This issue  can degrade    the model  performance    since the trained model is not familiar with the new target domain that has a large gap with the source domains. In the next section, we describe our test-time style shifting that handles this issue. However, in DG-specific imbalance scenarios with missing classes in each domain (i.e., specific classes missing in $S_n$), the gain of test-time style shifting could be limited. Therefore, to maximize the potential of test-time style shifting we also propose the style balancing strategy in the next section.

\section{Proposed Algorithm}
 \vspace{-0.5mm}

Section \ref{subsec:shift} describes our \textit{style balancing}, which  provides a good platform for our test-time style shifting by handling the  DG-specific imbalance issues during training. Based on the model obtained by our style balancing,   in Section \ref{subsec:projection}, we  propose  \textit{test-time style shifting} to handle the issue on the large style gap between source and target domains. A high-level description of our idea is shown in Fig. \ref{fig:algorithm}. 

\subsection{Style Balancing }\label{subsec:shift} 
\vspace{-0.5mm}

Style balancing   strategically  shifts  the  style (i.e., style statistics) of each train sample  to another source domain that has an insufficient number of   data samples for each class. Given a mini-batch, style balancing is applied to each class $k\in\{1,2,\dots, K\}$ independently. Hence,  we describe our scheme focusing on a specific   class $k$.


\textbf{Step 1: Determining the number  of samples to shift.}   We define $\tilde{S}_{n,k}$ as the set of  samples that belong to source domain $n$ labeled as class $k$ in a specific mini-batch. We would like to balance the number of samples across all source domains $n\in\{1,2,\dots,N\}$ so that each domain has average number of samples  $Q_k:= \frac{1}{N}\sum_{n=1}^N|\tilde{S}_{n,k}|$  for class $k$.  
If $|\tilde{S}_{n,k}| > Q_k$ holds, $|\tilde{S}_{n,k}| - Q_k$  samples in source domain $n$ should shift their styles to  other domains that have less than $Q_k$ samples. Otherwise (i.e., $|\tilde{S}_{n,k}| < Q_k$), we similarly shift the styles of samples in other domains (that have more than $Q_k$ samples) to domain $n$. 
Based on this,  one can easily determine the number of samples to be shifted from domain $n$ to another domain $n'$ for all $n,n'\in\{1,2,\dots,N\}$, in order to balance class $k$ across all source domains.


\textbf{Step 2: Sample selection.}  In this step, for domain $n$ satisfying $|\tilde{S}_{n,k}| >  Q_k$,  we strategically select $|\tilde{S}_{n,k}| -Q_k$ samples to be shifted from domain $n$ to other source domains. 
Our key insight is that samples having similar style statistics would provide similar effects on improving domain diversity, when existing DG schemes are applied. Based on this intuition, we propose to move the style of the sample that has   very similar style statistics with other samples.

We first define the distance between the style statistics of any two samples $s_i,s_j\in \tilde{S}_{n,k}$ as
\begin{equation}
d_{i,j} = \| \Phi(f(s_{i})) - \Phi(f(s_{j})) \|,
\end{equation}
where $\|\cdot\|$ is the Euclidean distance. Then we choose two samples $s_{i^*}$ and $s_{j^*}$  from $\tilde{S}_{n,k}$ that satisfy  $(i^*,j^*)=\text{argmin}_{(i,j)}d_{i,j}$;  these two samples have the closest style statistics so that  similar effect can be observed even when one of these samples is removed from source domain $n$. Among these two samples, we choose the sample that has a smaller minimum distance from other samples, and shift its style to another domain; we choose sample $s_{i^*}$ if $\text{min}\{d_{z, i^*}\}_{z=1, z\neq j^*}^{|\tilde{S}_{n,k}|} < \text{min}\{d_{z, j^*}\}_{z=1, z\neq i^*}^{|\tilde{S}_{n,k}|}$ and choose sample $s_{j^*}$, otherwise.      This process is repeated until $|\tilde{S}_{n,k}| - Q_k$ samples are selected from   domain $n$. We repeat this process for all source domains $n\in\{1,2,\dots,N\}$.


\textbf{Step 3: Balancing.} Suppose sample $s$ in domain $n$ has to shift its style    to   domain $n'$,    according to Steps 1 and   2 above.  We randomly select two samples $s_1', s_2'\in S_{n'}$ from  domain $n'$  and shift the style of $s$ to $s_1'$, $s_2'$ via EFDM, and apply EFDMix. Specifically, our style balancing (SB) performs
\begin{equation}\label{eq:SBmixing}
\text{SB}(f(s))_{\tau_i} = \lambda  f(s_1')_{\kappa_i} + (1-\lambda)f(s_2')_{\eta_i} + f(s)_{\tau_i}  - \langle f(s)_{\tau_i}  \rangle,
\end{equation}
where $\tau_i$, $\kappa_i$, $\eta_i$ are the indices of the  $i$-th smallest elements of vectors $f(s)$, $f(s_1')$, $f(s_2')$, respectively. $\langle\cdot\rangle$ is the stop gradient operation; $\langle f(s) \rangle$ is the copy of $f(s)$ detached from computational graph. The term $f(s) - \langle f(s) \rangle$ is introduced to facilitate backpropagation of sample $s$ as in \cite{zhang2022exact}. The process in (\ref{eq:SBmixing}) eventually shifts the style of sample $s$ in source domain $n$ to    domain $n'$.    $\lambda$ is a mixing parameter which is sampled from the Beta distribution.  

The above three steps are applied to the samples in each class $k\in\{1,2,\dots,K\}$ independently.  By balancing the number of samples  for each class across all source domains, our style balancing  not only handles the DG-specific imbalance issues but also maximizes the advantage  of  test-time style shifting, as described in the next subsection. 

\subsection{Test-Time Style Shifting}\label{subsec:projection} 
\vspace{-0.5mm}


Our test-time style shifting   strategy shifts the styles of the test samples  during testing to handle arbitrary target domains. If the test sample has a large style gap with the source domains,  then the style of the test sample is shifted to the nearest source domain  that the model is already familiar with, before making the prediction. Otherwise,  the test sample keeps its original style.

Let $t\in T$ be the test sample  from an arbitrary unseen domain in test set $T$, where $f(t)$ is the encoded features of $t$ at a specific layer. Recall that $\mu(f(t))$ and $\sigma(f(t))$ are the channel-wise mean and standard deviation of $f(t)$. Also recall that the mean of feature statistics in each source domain $n\in\{1,2,\dots,N\}$ are written as $\mu_{S_n} = \frac{1}{|S_n|}\sum_{s\in S_n}\mu(f(s))$ and $\sigma_{S_n} = \frac{1}{|S_n|}\sum_{s\in S_n}\sigma(f(s))$.  We also define the mean feature statistics averaged over all source domains as $\mu_S = \frac{1}{N}\sum_{n=1}^N \mu_{S_n}$ and  $\sigma_S = \frac{1}{N}\sum_{n=1}^N \sigma_{S_n}$.  According to the definition in (\ref{eq:defn}), we have $\Phi_{S_n} = [\mu_{S_n}, \sigma_{S_n}]$, $\Phi_S = [\mu_S, \sigma_S]$.

Based on these notations, at a specific layer, we generate new style statistics of sample $t$  as $\Phi(f(t))_{\text{new}}=$
{\small\begin{equation} \label{eq:ttss_algorithm}
 \begin{dcases}
\Phi_{S_{n'}}  \  \ \text{if} \ \  \frac{1}{N}\sum_{n=1}^N \|\Phi(f(t)) - \Phi_{S_n} \|   >  \alpha \Big( \frac{1}{N}\sum_{n=1}^N \|\Phi_S - \Phi_{S_n} \| \Big) \\
\Phi(f(t))  \  \  \    \text{otherwise},
\end{dcases}
\end{equation}}
where $\Phi(f(t))_{\text{new}} = [\mu(f(t))_{\text{new}},\sigma(f(t))_{\text{new}}]$, $n'$ is the index of the closest source domain  to the test sample $t$, i.e., $n'=\text{argmin}_n\|\Phi(f(t)) - \Phi_{S_n}\|$, and  $\alpha$ is a hyperparameter greater than or equal to 0.  

Now based on   $\mu(f(t))_{\text{new}}$ and $\sigma(f(t))_{\text{new}}$, following the process of AdaIN in (\ref{eq:adain}), our test-time style shifting (TS) shifts the style of sample $t$ while preserving its content  as
\begin{equation}
\text{TS}(f(t)) = \sigma(f(t))_{\text{new}}\frac{f(t)-\mu(f(t))}{\sigma(f(t))} + \mu(f(t))_{\text{new}}.
\end{equation}
\textbf{Intuitions.} In (\ref{eq:ttss_algorithm}), if there is a large gap between style statistics of source domains and the test sample, we  shift  the style statistics of the test  sample to the \textit{nearest source domain}. This enables  predictions on the domain that  the model is already familiar with. Otherwise,  i.e., when the style gap   is acceptable, the model is likely to be well-trained on the style of the test sample. Thus, we let the test  sample $t$  keep its  current  style.  This strategy enables the model to handle any target domains with arbitrary styles. Moreover, compared to the existing test-time adaptation ideas, our scheme requires less computational burden at testing since only AdaIN is required without any model update process.

\textbf{Remark.} Consider a DG-specific imbalance scenario where some of the classes are missing in each domain. When test-time style shifting is applied without performing style balancing (of Section  \ref{subsec:shift}),  the model performance could be limited since the  trained model does not make reliable predictions even for the samples in the source domains.     Hence, it is advantageous to perform style balancing during training to improve the effectiveness of test-time style shifting.  

 \subsection{Overall Procedure and Discussions}\label{subsec:overall}
\vspace{-0.5mm}

The overall procedure of our algorithm is shown in Fig.  \ref{fig:algorithm}. Given imbalanced style  statistics, we first perform style balancing. Then, we can apply any DG methods for training (e.g., style augmentation). 
When training is finished, we apply our  test-time style shifting   
and make a prediction.

\textbf{Where to apply SB and TS.} Our style balancing (SB) and test-time style shifting (TS) can be flexibly applied at any layer  of the backbone. During training, we only have the SB module, which is discarded when training is finished. During testing, the TS module is applied at a predetermined layer. Various ablations and of  SB/TS modules are provided in Section \ref{sec:experiment} and Appendix.

\textbf{Compatibility with  various DG methods.}  The simplest way to combine our work with others is to apply SB  before style augmentation  (e.g., MixStyle), which also work in the style space as our scheme.  Due to the high flexibility of  SB and TS modules,  our method can also work in conjunction with other DG strategies.    
For example,  the  SB module can be applied at the inner optimization process of meta-learning DG approach  \cite{li2018learning} to handle the imbalance issues in the meta-train source domains.  As another example,  our SB can be applied at   the feature learning network of  conditional invariant deep DG method \cite{li2018deep}. For all methods,  TS can be applied at a specific layer of the network  during testing. In Section  \ref{sec:experiment}, we show   that  SB and TS are compatible not only with  style augmentation based schemes but also with other DG methods relying on meta-learning or domain alignment. 

\textbf{Hyperparameters.} In our SB,  the  mixing parameter  $\lambda$ in (\ref{eq:SBmixing}) is sampled from Beta distribution as $\lambda \sim Beta(\tau, \tau)$. This parameter also appears in MixStyle \cite{zhou2021domain} and EFDMix \cite{zhang2022exact}, and we set $\tau=0.1$  for all experiments as in these prior works. Compared to existing style augmentation methods, our scheme requires an  additional hyperparameter   $\alpha$   that appears in  (\ref{eq:ttss_algorithm}) of our TS module, which is set to 3 for all classification results.  A  detailed discussion regarding $\alpha$ is provided  in Appendix. 

\textbf{Complexity.} 
Once the style statistics of train samples are obtained,  only the style gaps between the test sample and the center of $N$ source domains are required for  test-time style shifting; this  makes the additional complexity  negligible  compared with existing test-time adaptation methods that require additional model updates. Our strategy only require AdaIN  during testing.  Regarding style balancing,  suppose that there are $\frac{B}{NK}$ samples in a mini-batch corresponding to each domain $n$ with class label $k$. Then, the additional complexity required for our style balancing (during training) becomes $\mathcal{O}((\frac{B}{NK})^2 \times N \times K) = \mathcal{O}(\frac{B^2}{NK})$, which is the additional cost for achieving an improved domain diversity.

\section{Experimental Results}\label{sec:experiment}

 
\subsection{Generalization on Multi-Domain Classification} \label{subsec:51}
\textbf{Experimental setup.}  Targeting multi-domain classification, we perform experiments using PACS  \cite{li2017deeper} with 4 domains (Art, Cartoon, Photo, Sketch) and  VLCS \cite{fang2013unbiased} with 4 domains (Caltech, LabelMe, Pascal, Sun),   
which are the commonly adopted benchmarks for DG. We also considered Office-Home \cite{venkateswara2017deep} dataset in Appendix.  We  focus on the leave-one-domain-out setting where the model is trained on three domains  and tested on the remaining one domain. The case with single-domain generalization is considered in   Section \ref{sec:further}.  Following the setups in \cite{zhou2021domain, li2022uncertainty, zhang2022exact}, we adopt ResNet-18  pre-trained on ImageNet as a backbone, where the results with ResNet-50 are reported in Appendix. For PACS, the proposed SB module is probabilistically operated once at first or second or third residual blocks during training, while the TS module  is operated at the second residual block during testing. Other implementation details and ablations on SB/TS  locations are provided in Appendix.   We utilize the term ``TSB'' for the scheme that uses TS and SB simultaneously. 
\vspace{-0.5mm}

We consider not only the original PACS and VLCS   but also  the imbalanced  version of each dataset. We consider two different imbalance scenarios: cross-domain data imbalance and cross-domain class imbalance scenarios. To model the first scenario, we keep the training data of the largest source domain while  removing  a specific portion of training data of the remaining two source domains, which will be clarified soon. When constructing the cross-domain class-imbalanced dataset, among 7 classes in PACS, we select 3 classes from the first source domain, other 2 classes from the second source domain, and the remaining 2 classes from the last source domain. 
In VLCS, among 5 classes, we select 2, 2, 1 classes from each source domain to construct the imbalanced dataset.  This effectively models the missing classes in each domain. The class imbalanced dataset could be also constructed in different settings, e.g., in a long-tailed imbalance setting \cite{cao2019learning}. The corresponding results are reported in Appendix.  The performance is obtained by averaging the results over 5 independent trials.  More details on our experimental setup  are   provided in Appendix. 

\vspace{-0.5mm}

\begin{table}[t]
\vspace{-2mm}
	\tiny
	\caption{\small  Results on \textbf{original PACS}. We reproduced the results of MixStyle, DSU, EFDMix while other   values are from  original papers (denoted with *). TS plays a key role in improving the  model  accuracy on original PACS, especially on the Sketch domain that has a large style gap with other domains (as shown in Fig. \ref{fig:tsne}).  }
	\centering
	\label{tab:original_pacs}
	\begin{tabular}{l|    cccc | c}
		\toprule  
		Methods   &Art  & Cartoon & Photo & Sketch & Avg. \\ 
		\midrule
		$\text{L2A-OT}^*$ \cite{zhou2020learning} & 83.3& 78.2 & 96.2  & 73.6  &82.8 \\
		 
		$\text{pAdaIN}^*$  \cite{nuriel2021permuted}&   81.74 & 76.91  & 96.29  & 75.13 & 82.51\\
		
		$\text{SagNet}^*$ \cite{nam2021reducing}&   83.58 & 77.66  & 95.47  & 76.3 & 83.25\\
		$\text{Tent}^*$ \cite{wang2020tent}&  81.55& 77.67& 95.49  & 77.64 & 83.09\\
		$\text{T3A}^*$ \cite{iwasawa2021test}& 80.4 & 75.2 & 94.7 & 76.5 & 81.7 \\
		$\text{SSG}^*$ \cite{xiao2022learning}&   82.02 & 79.73  & 95.87  & 78.96 & 84.15\\
		\midrule
		Baseline - ResNet18&    73.97&	74.71&	96.07	&65.71&	77.62
 \\
		SB (Baseline) &     80.55&	77.16&	96.39&	71.68	&81.44 \\
		TS (Baseline)&   73.89&	75.14&	95.87 &	72.00	& 79.23 \\
		\textbf{TSB} (Baseline)  & 80.60	&77.58&	96.35&	74.37&	\underline{\textbf{82.22}}

\\
				\midrule		
		MixStyle  \cite{zhou2021domain} &  82.54 & 79.42 & 95.88 & 74.06 & 82.98 \\
		SB  (+ MixStyle) & 83.48 & 79.07&96.15&73.74&83.11  \\
		 TS (+ MixStyle)     & 82.59 & 79.99& 95.88& 78.66&84.28\\
		\textbf{TSB} (+ MixStyle)  & 83.62&80.07&96.15&78.66&\underline{\textbf{84.63}}\\
				\midrule
		DSU  \cite{li2022uncertainty}  &  81.78 & 78.66 & 95.91 & 76.75 & 83.27 \\
		SB  (+ DSU)&80.98&79.61&95.95&78.66&83.80 \\
	 TS  (+ DSU) & 81.12 & 80.31 & 95.82 & 79.19&84.11   \\
		\textbf{TSB} (+ DSU)    & 80.73 & 80.69 & 95.83& 79.47 &\underline{\textbf{84.18}}\\
								\midrule
		EFDMix  \cite{zhang2022exact}   & 83.12 & 79.76 & 96.43 & 75.08 & 83.60  \\
		SB  (+ EFDMix) & 83.98 & 79.75&96.47&75.12&83.83\\
	 TS (+ EFDMix)   & 83.05&81.31&96.40&78.93&84.92  \\
		\textbf{TSB} (+ EFDMix)   & 84.00 & 80.72 & 96.46 & 78.85 & \underline{\textbf{85.00}} \\
		\bottomrule
	\end{tabular}
	\vspace{-2mm}
\end{table}

 \begin{figure}[t]
\centering
  \begin{subfigure}[b]{0.235\textwidth}
  \centering
           \includegraphics[width=\textwidth]{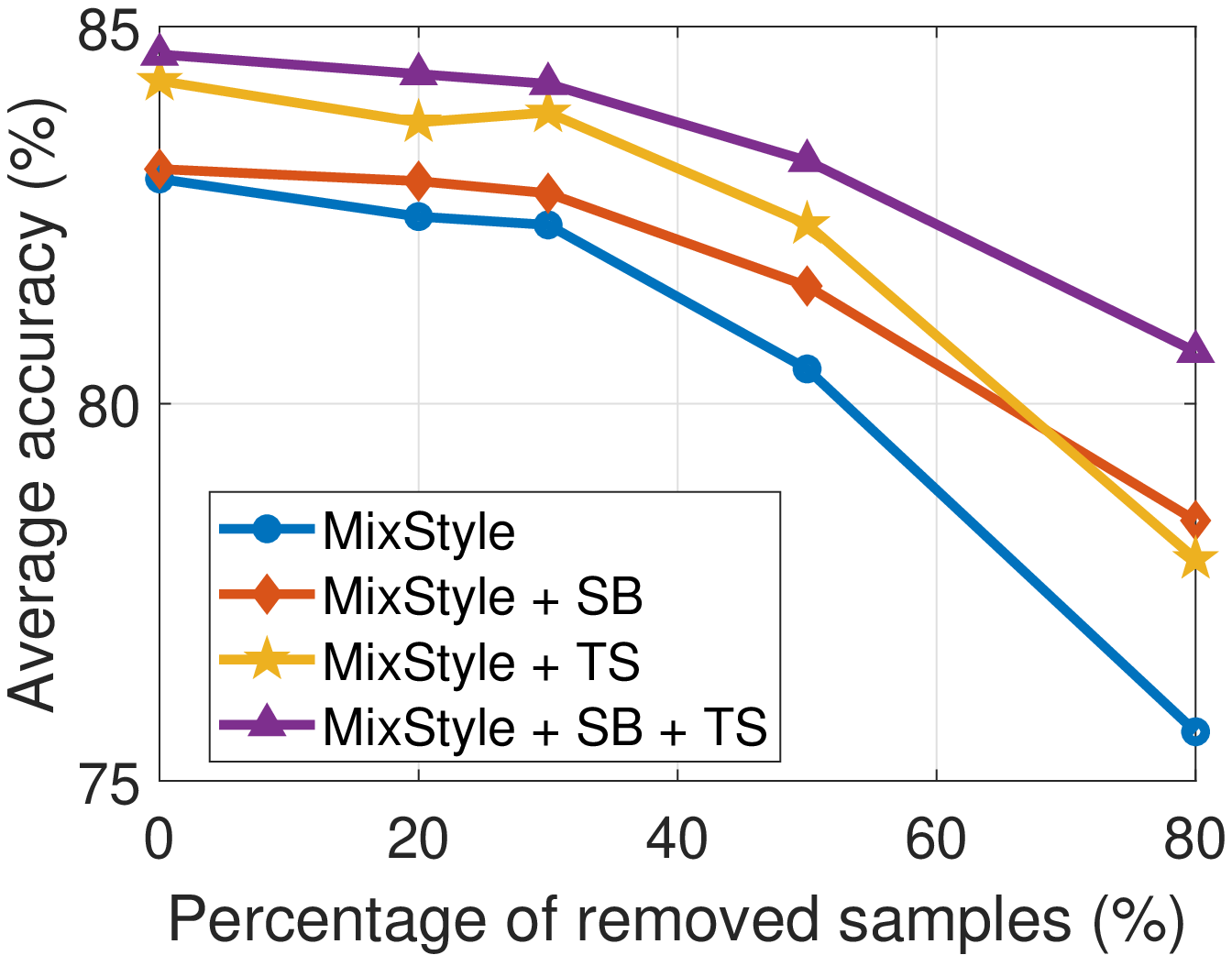}
  \caption{Effect  of SB/TS on MixStyle}
  \end{subfigure}
  \hfill
  \begin{subfigure}[b]{0.235\textwidth} 
  \centering
           \includegraphics[width=\textwidth]{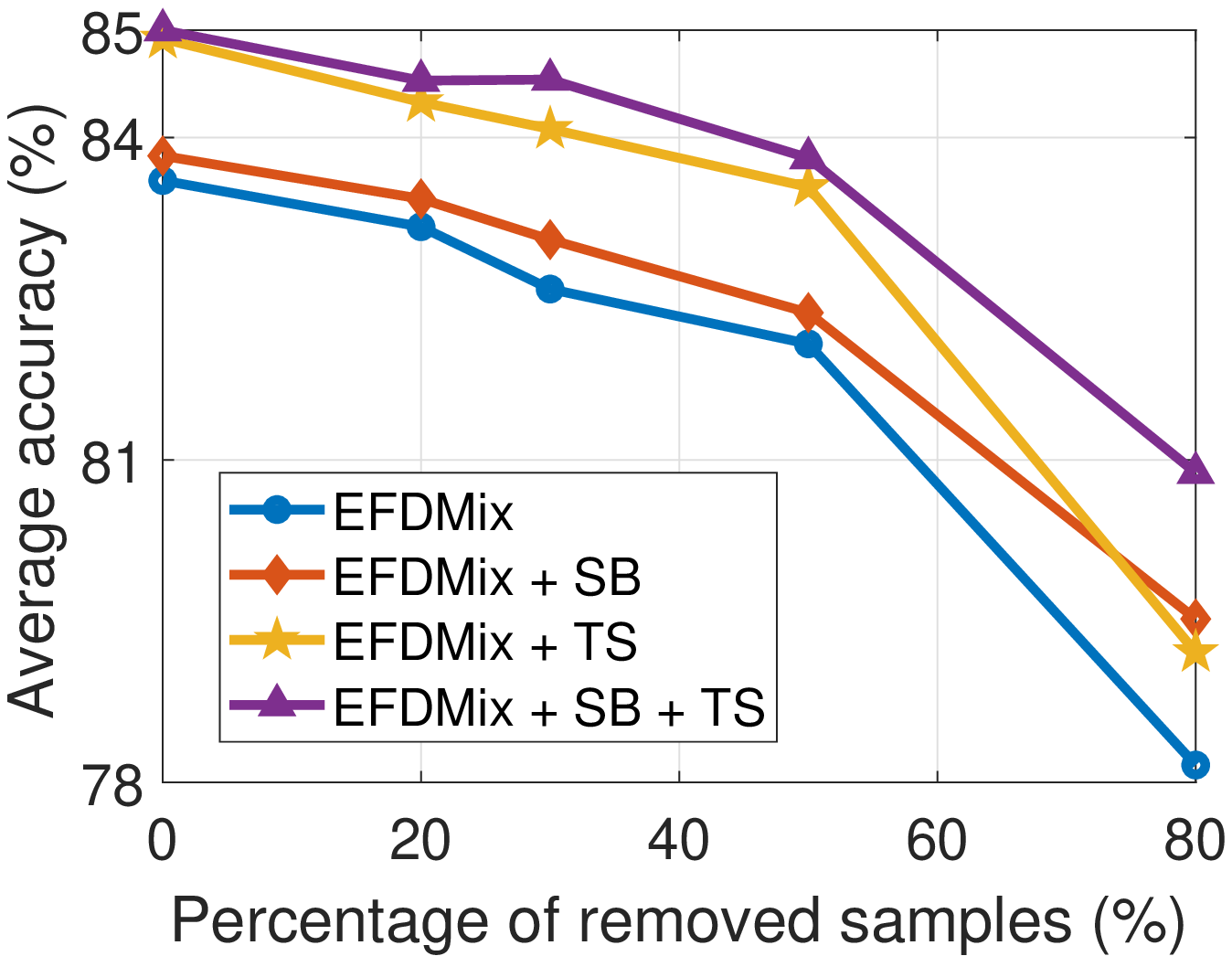}
  \caption{Effect of SB/TS on EFDMix}
  \end{subfigure}        
  \vspace{-1.5mm}
  \caption{ \small Results on \textbf{domain-imbalanced PACS}. We remove the samples of each source domains except the largest one. Both SB and TS are effective in improving the model performance.}
  \vspace{-1mm}
    \label{fig:varying_remove}
\end{figure}
 
  \begin{table*}[t]
    \centering
	\footnotesize
	\caption{\small Results on \textbf{imbalanced PACS}.  Compared to   Table \ref{tab:original_pacs},  the role of SB becomes more significant in severely imbalanced scenarios; SB not only improves the model performance by itself but also provides a good platform for maximizing the advantage of TS.  }
	\centering
	\label{tab:imbalanced_pacs}
	\begin{tabular}{l|  c | cccc | c  ||  cccc | c    }
		\toprule  
		\multirow{2}{*}{Methods} & \multirow{2}{*}{Reference}&  \multicolumn{5}{c||}{\textbf{Cross-domain data  imbalance}}  & \multicolumn{5}{c}{\textbf{Cross-domain class imbalance}} \\
	 &  &	Art  & Cartoon & Photo & Sketch & Avg. &Art  & Cartoon & Photo & Sketch &  Avg.   \\
		\midrule
		MixStyle  & ICLR'21&71.73&	73.80&	90.60&	66.48	&75.65&  39.91	&54.08&	56.45	&44.82	&48.82 
\\
		 SB (+ MixStyle) &   Ours   & 76.53	& 75.61 &	93.33	&68.34	&78.45&44.49&	55.57&	56.28&	44.93&	50.32
 \\
		 TS  (+ MixStyle)&  Ours & 72.04&	74.01	&90.60&	75.12&	77.94& 39.98&	54.01&	56.45	&44.44&	48.74
\\
		\textbf{TSB} (+ MixStyle) & Ours & 76.97	&76.62	&93.29	&75.88&	\underline{\textbf{80.69}}& 44.50	& 55.84	& 56.28& 	46.68	& \underline{\textbf{50.83}} 
\\

				\midrule
		DSU  & ICLR'22& 75.76	& 75.26	&91.90&	72.45	&78.84&  29.61	&45.24	&46.90	&39.37	&40.28 
\\
		SB (+ DSU)&  Ours  & 76.04	&76.15	&92.87&	73.47	& 79.64&45.09&	53.93	&60.25	&47.74&	51.75
\\
		TS (+ DSU) &  Ours & 75.49	&76.69&	91.92&	76.36&	80.12 &  29.78	&44.54	&46.90&	36.65	&39.47 
 \\
		\textbf{TSB}  (+ DSU)& Ours & 75.93& 	77.39	& 92.85& 	75.90	& \underline{\textbf{80.52}}& 45.03&	54.42&	60.24&	49.20&	\underline{\textbf{52.22}} 
\\
						\midrule
		EFDMix   &CVPR'22 & 75.33&	75.67	&90.59&	71.07	&78.16& 44.68&	54.87&	58.15&	44.64	&50.59
 \\
		 SB (+ EFDMix)&  Ours  & 77.91	& 76.38&	92.79&	70.99	&79.52 & 46.63&	54.84	&57.89	&44.47&	50.96

\\
	 TS (+ EFDMix) & Ours & 75.39&	75.92	&90.56	&74.97&	79.21 &44.56&	55.05&	58.15&	45.96&	50.93\\
		\textbf{TSB} (+ EFDMix)& Ours & 77.90&	76.54	&92.71&	76.37	&\underline{\textbf{80.88}}& 46.03&	55.29 &	57.87&	49.99&	\underline{\textbf{52.30}} 
\\
		\bottomrule
	\end{tabular}
			\vspace{-2mm}
\end{table*}

\textbf{Baselines.}   First, we consider  the state-of-the-art  style  augmentation schemes,   MixStyle \cite{zhou2021domain}, DSU \cite{li2022uncertainty}, EFDMix \cite{zhang2022exact}, that also work in the style space as ours.  We apply our SB and TS to these schemes to validate the effectiveness of the proposed ideas. To confirm the compatibility with other DG methods, we also apply our SB/TS to  MLDG  \cite{li2018learning} and    CDANN \cite{li2018deep}   in Section \ref{sec:further}.  For a fair comparison, all hyperparemters are set to be same as in the original setup of each baseline. We also apply our methods to the  pure baseline without any DG algorithm.   The following other recent works are also considered: L2A-OT \cite{zhou2020learning}, pAdaIN \cite{nuriel2021permuted}, SagNet \cite{nam2021reducing}. We also compare our scheme with the recent test-time adaptation works: Single sample generalization (SSG) \cite{xiao2022learning}, T3A \cite{iwasawa2021test} and Tent \cite{wang2020tent}. A more detailed comparison with the test-time adaptation works is reported in Appendix. Finally, in Appendix, we compare our SB module with  the recent work BoDA  \cite{yang2022multi}  that tackles the imbalance issues in a multi-domain setup.  

\textbf{Result 1: Original dataset.} We first observe Table \ref{tab:original_pacs}, which shows the results on original PACS. Both  SB  and TS  play important roles in   all baselines. The performance gain of SB is noticeable since PACS is already slightly imbalanced across domains.   The performance gain of TS is especially large  in Sketch, since  the Sketch domain  has a large style gap with other source domains (see Fig. \ref{fig:tsne}).   The overall results show that our scheme significantly boosts up the performance of recent style-augmentation  methods. Our scheme also   outperforms other recent methods for  DG.

\begin{table}[t]
 \vspace{-1mm}
	\scriptsize
	\caption{\small  Results on  \textbf{imbalanced VLCS}. }
	\centering
	\label{tab:imbalanced_vlcs} 
	\begin{tabular}{l|   cccc | c}
		\toprule  
		Methods   & Caltech   & LabelMe & Pascal& Sun & Avg. \\
		\midrule
		
		
		
		MixStyle   &  68.87	&53.32	&55.12	&39.09	&54.10\\
		SB (+ MixStyle) &  69.97&	53.87	&55.51	&38.51&	54.47\\
		\textbf{TS} (+ MixStyle)    &  73.51&	53.20	&55.15	&38.98	& \underline{\textbf{55.21}}\\
		 TSB (+ MixStyle)    & 73.27&	53.78	&55.02&	38.58&	55.16\\
				\midrule
		DSU  &   63.07	& 54.13& 	56.01	& 39.90& 	53.28
 \\
		SB (+ DSU)&   74.02&	53.40&	55.91	&40.22	&55.89\\
		TS (+ DSU)  &     65.99	&53.90	&55.93	&40.02& 53.96\\
		 \textbf{TSB} (+ DSU)     & 75.99	&53.50	&55.46&	40.28&	\underline{\textbf{56.31}}\\
	
		\bottomrule
	\end{tabular}
				\vspace{-2mm}
\end{table}

\textbf{Result 2: Cross-domain data imbalanced dataset.}  In Fig. \ref{fig:varying_remove}, we plot the average accuracy on the imbalanced version of PACS. We removed a certain portion of all source domains except the largest one. It can be observed that the gain of SB becomes larger as the portion of removed samples increases, i.e., as the dataset becomes more severely imbalanced. The TS module is effective for all cases.   The left part of Table \ref{tab:imbalanced_pacs} shows the full result for the case of $80\%$.  The advantage of SB is significant compared to the case in Table \ref{tab:original_pacs}; the major performance gains of Art, Cartoon, Photo come from SB,   showing the effectiveness of SB to improve the domain diversity in imbalanced datasets. On the other hand, the main performance gain of Sketch comes from TS as in Table  \ref{tab:original_pacs};  again, this is because Sketch has a significant style gap with other  source domains as shown in   Fig. \ref{fig:tsne}.  The overall results confirm the advantage of both SB and TS.

\textbf{Result 3: Cross-domain class imbalanced dataset.} In cross-domain class imbalance scenario (right part of Table \ref{tab:imbalanced_pacs}), different from the trends in original dataset and cross-domain data imbalanced dataset, directly applying TS (without SB) does not improve the performance in general (even in Sketch). This is because the model trained without SB lack generalization capability in this scenario,  indicating that SB also plays a key role for maximizing the advantage of TS. The performance gain of SB is especially large when combined with DSU;  compared to MixStyle or EFDMix, in DSU, each class tends to get exposed to only the styles that are close to the original source domain without SB. 
 Table \ref{tab:imbalanced_vlcs} shows the performance on cross-domain class imbalanced VLCS. Although the performance gain is smaller compared to PACS due to the small style gaps of source and target domains, the trend is consistent with the results in PACS.

\begin{table*}[t] 
	\footnotesize
	\caption{\small  Results on \textbf{person re-ID task}, using  Market1501 and GRID datasets. }
	\centering
	\label{tab:retrieval}
	\begin{tabular}{l|  c | cccc|c c c  c}
		\toprule  
		\multirow{2}{*}{Methods} & \multirow{2}{*}{Reference} &  \multicolumn{4}{c|}{Market $\rightarrow$  GRID} &  \multicolumn{4}{c}{GRID $\rightarrow$  Market} \\
		 &  & mAP & R1 & R5 &R10& mAP & R1 & R5 &R10\\
		\midrule
	 	MixStyle  \cite{zhou2021domain}& ICLR'21& 35.30	&26.67	&\textbf{44.53}	&53.07 & 5.25	&16.40	&30.05&	37.05\\
		\textbf{TSB} (+ MixStyle) & Ours& \textbf{36.30}&	\textbf{28.27}&	42.93&	\textbf{55.47} &\textbf{5.70}	&\textbf{17.75}&	\textbf{31.90}&	\textbf{39.65}\\
		\midrule
		DSU  \cite{li2022uncertainty}& ICLR'22 &38.57	&30.40	&46.40	&53.07 & 4.45	&14.90&	27.65&	34.60 \\
		\textbf{TSB} (+ DSU)  & Ours & \textbf{40.10}&	\textbf{30.67}&	\textbf{48.00}	&\textbf{58.13} & \textbf{5.25}	&\textbf{16.70}	&\textbf{31.60}&	\textbf{38.85}\\
		\midrule
		EFDMix \cite{zhang2022exact}& CVPR'22 &36.33	&\textbf{27.47}	&45.87	&52.27 & 6.07	&19.27&	33.70&41.30\\
		\textbf{TSB} (+ EFDMix)  & Ours & \textbf{36.67}&	26.93&	\textbf{46.67}	&\textbf{55.57} & \textbf{6.53}	&\textbf{20.23}	&\textbf{35.37}&	\textbf{43.13}\\



		\bottomrule
	\end{tabular}
		\vspace{-3mm}
\end{table*}


		\vspace{-1mm}

\subsection{Generalization on Instance Retrieval} \label{subsec:53}
\vspace{-0.5mm}

We also consider a different task, known as multi-domain instance retrieval. We   consider person re-identification (re-ID), where the goal is to match the same person   using various camera views. This setup can be viewed as a multi-domain image matching problem  by regarding different camera views as distinct domains.  As in  the setup of \cite{zhang2022exact}, we adopt Market1501 \cite{zheng2015scalable} and GRID \cite{loy2009multi} datasets, and train the model in one dataset and test on the other one.  We train OSNet \cite{zhou2019omni} which was specifically designed for person re-ID. Other details are provided in Appendix. Table \ref{tab:retrieval} shows the corresponding results, indicating that our idea is powerful even in multi-domain image matching problem.


\vspace{-1mm}

\subsection{Further Experiments and Discussions}\label{sec:further}
\vspace{-0.5mm}

\textbf{Compatibility with other DG methods.} 
Due to the flexibility of our SB and TS modules, our scheme can also work effectively with other DG strategies based on meta-learning and domain alignment. 
\begin{table}
	\scriptsize
	\caption{\small  Compatibility   with other DG strategies.}
	\centering 
	\label{tab:otherDG}
	\begin{tabular}{l|     cccc | c}
		\toprule  
		Methods  & Art &Cartoon & Photo& Sketch &  Avg. \\ 
		\midrule  
		MLDG  &31.18  &46.70 &44.76 & 38.39 &  	40.26\\
		 SB	(+ MLDG) & 42.89& 60.66& 58.68& 50.89 & 53.28 \\
		\textbf{TSB} (+ MLDG) &42.95& 61.67& 58.68& 50.86  & \underline{\textbf{53.54}} \\
		\midrule
		CDANN   &17.69& 24.36& 27.25& 33.14  & 25.61 	\\
		   SB	(+ 	CDANN) & 40.94& 52.61& 50.15& 41.16 & 46.21 \\
		\textbf{TSB} (+ 	CDANN) & 40.94& 52.61& 50.15& 41.16 & \underline{\textbf{46.22}}   \\ 
	\bottomrule
	\end{tabular}
	\vspace{-3mm}  
\end{table}
Table \ref{tab:otherDG} shows the results of MLDG \cite{li2018learning} (meta-learning based method) and CDANN \cite{li2018deep}  (domain alignment based method) combined with our scheme on cross-domain class imbalanced PACS. We consider the DomainBed setup for experiments. It can be seen  that our scheme improves the model performance of both methods, confirming that both SB and TS can work in conjunction with various DG methods to mitigate the  cross-domain imbalance issue and the style gap issue concurrently.

\begin{table}
	\footnotesize
	\caption{\small  Comparing SB with existing class imbalanced  methods. } 
	\centering 
	\label{tab:classimbalance} 
	\begin{tabular}{l|     c}
		\toprule  
		Methods  &  Avg. accuracy \\ 
		\midrule 
		DSU  \cite{li2022uncertainty}  &  40.28  \\
		\textbf{SB} (+ DSU)    & \textbf{51.75}  \\
		\midrule 
		DSU  + Undersampling  	& 40.37 \\
		\textbf{SB} (+ DSU) + Undersampling  	&  \textbf{47.74}\\
				\midrule 
		DSU +  Oversampling  & 43.91  \\
		\textbf{SB} (+ DSU) +   Oversampling   	&   \textbf{54.01}  \\
				\midrule 
	DSU  + Reweighting  & 41.57   \\
	\textbf{SB} (+ DSU) +   Reweighting   &  \textbf{52.57} \\
	\bottomrule
	\end{tabular}
	\vspace{-3mm}  
\end{table}

\textbf{Comparing SB with existing class imbalanced  methods.} 
In Table \ref{tab:classimbalance}, we   compare   SB  with  existing class imbalanced learning methods in a cross-domain class imbalance scenario, under the same setup in Table \ref{tab:imbalanced_pacs}. We consider  
the following baselines:  undersampling majority classes,  oversampling minority classes,  reweighting the objective function based on the \textit{effective number}   \cite{cui2019class}.   It can be seen that     existing methods generally fail to handle the  issue since the missing classes in each domain  cannot be compensated via over/under-sampling or reweighting in this DG-specific imbalance setup. 
Our SB  effectively alleviates this  issue, significantly improving the model performance.

\begin{table}
	\scriptsize
	\caption{\small  Results in a single-domain generalization setup. }
	\centering
	\label{tab:single_domain}
	\begin{tabular}{l|   cccc | c}
		\toprule  
		Methods   & Art   & Cartoon & Photo & Sketch & Avg. \\
		\midrule		
		MixStyle   &  64.32 &71.77& 42.98& 32.18& 52.81 \\
		\textbf{TS} (+ MixStyle)  & 72.19& 77.25& 48.50& 43.62& \underline{\textbf{60.39}} \\
				\midrule
		DSU   &  64.85& 74.53& 39.48& 36.20& 53.77 \\
		\textbf{TS }  (+ DSU)&   70.99& 73.95& 51.18& 49.03& \underline{\textbf{61.28}} \\
				\midrule
	 	EFDMix    &  66.56& 73.93& 44.74& 36.36& 55.40    \\
	 	\textbf{TS}   (+ EFDMix)  & 73.87& 76.79& 53.04& 49.41& \underline{\textbf{63.28}}  \\	
		\bottomrule
	\end{tabular}
				\vspace{-3mm}
\end{table}  

 \textbf{Single-domain generalization.} In Table \ref{tab:single_domain}, we also show the results in a single-domain generalization setup using the original PACS dataset: the model is trained with one source domain, and tested on the remaining three domains to measure the model accuracy. We do not apply SB since only one source domain exists during training.  At testing, we shift the style statistics of all test samples to the source domain.  It can be seen that our TS   significantly boosts up the performance of existing methods   in a single domain setup by simply shifting the style statistics of the test samples.

\textbf{Additional experimental results.} Other results including  results in a DomainBed setup, detailed comparison with test-time adaptation works,  results without domain labels,   results on long-tailed imbalance settings, and results on the Office-Home  dataset are shown in Appendix. We also perform  additional studies on our SB/TS modules, in Appendix.





	\vspace{-1mm}   

\section{Conclusion} 
In this paper, we proposed test-time style shifting, a simple yet effective strategy that can handle the domain shift issue in DG. By shifting the styles of specific test samples to the nearest  the source domain before making predictions, our scheme is able to handle any target domains with arbitrary style statistics. We also proposed style balancing, to increase the potential of test-time style shifting while handling the DG-specific imbalance issues. Experimental results on various datasets and data distribution scenarios confirmed the effectiveness of the proposed ideas, providing new guidelines for DG in practice with imbalance and domain shift issues.

\section{Acknowledgments}

This work was supported by the National Research Foundation of Korea (NRF) grant funded by the Korea government (MSIT) (No. NRF-2019R1I1A2A02061135), and by Center for Applied Research in Artificial Intelligence (CARAI) grant funded by DAPA and ADD (UD230017TD), and  by IITP funds from MSIT of Korea
(No. 2020-0-00626).

\nocite{langley00}

\bibliography{egbib1}
\bibliographystyle{icml2023}

\newpage
\appendix
\onecolumn
\section{Comparison with other State-of-the-Arts in DomainBed Setup}
Following the DomainBed setup \cite{gulrajani2020search}, in Table \ref{tab:domainbed}, we compare our approach with other state-of-the-arts using ResNet-50. Training-domain validation strategy is used for selecting the model in DomainBed setup. It can be seen that the proposed scheme combined with MixStyle achieves the best performance with average accuracy of $86.6\%$. We also combine our scheme with one of the state-of-the-art benchmarks, termed SWAD \cite{cha2021swad}. It is shown that our scheme can further improve the performance of the existing method. The overall results in Table \ref{tab:domainbed} show that our style balancing (SB) and test-time style shifting (TS) can be easily combined with other state-of-the-arts to achieve the best performance. 
\begin{table*}[ht]
	\vspace{-2mm}
	\caption{\small Performance in DomainBed setup. }
	\centering
	\label{tab:domainbed}
	\begin{tabular}{l|  c  ccc| c }
		\toprule  
		Methods  &Art  & Cartoon & Photo & Sketch & Average \\
				\midrule		
		ERM \cite{vapnik1999overview}& 84.7 &   80.8 &  97.2 &  79.3 & 85.5 \\
		IRM \cite{arjovsky2019invariant} & 84.8 & 76.4  & 96.7  & 76.1&  83.5  \\
		GroupDRO \cite{sagawa2019distributionally} &   83.5& 79.1 & 96.7 & 78.3& 84.4   \\
		Mixup \cite{zhang2017mixup} &   86.1 & 78.9 &  97.6 & 75.8 & 84.6 \\
		MLDG \cite{li2018learning}&  85.5 & 80.1 & 97.4&76.6 &84.9  \\
		CORAL \cite{sun2016deep} &  88.3 & 80.0& 97.5 & 78.8& 86.2 \\
		MMD \cite{li2018deep} & 86.1&79.4 &96.6 &76.5 &84.6  \\
		DANN \cite{ganin2016domain} &  86.4&77.4 &97.3 & 73.5 & 83.6  \\
		CDANN \cite{li2018deep} & 84.6 & 75.5& 96.8 & 73.5 &82.6  \\
		MTL \cite{blanchard2017domain} &   87.5&77.1 & 96.4 &77.3 &84.6 \\
		SagNet \cite{nam2021reducing} &  87.4 & 80.7 & 97.1 & 80.0 & 86.3  \\
		ARM \cite{zhang2020adaptive} &  86.8 & 76.8 & 97.4 & 79.3 & 85.1  \\
		VREx \cite{krueger2021out} &86.0 &79.1 & 96.9 & 77.7 & 84.9   \\
		RSC \cite{huang2020self}& 85.4 &79.7& 97.6&78.2 & 85.2  \\
		EFDMix  \cite{zhang2022exact},& 86.7& 80.3 & 96.3 & 80.8 & 86.0  \\
		MixStyle \cite{zhou2021domain}& 85.6 & 80.6 & 95.5 & 81.6  & 85.8  \\
		\textbf{SB (ours)} (+MixStyle) &  87.8 & 82.1  & 95.6 & 81.0  & \underline{\textbf{86.6}}   \\
	\midrule	
  \multicolumn{1}{c}{\hspace{-17mm}\textit{Combination with SWAD} 	} \\ 
 \midrule
SWAD  \cite{cha2021swad}	 &  89.3 &  83.4& \underline{\textbf{97.3}}& 82.5& 88.1 \\
	SWAD + MixStyle & 90.3 & 84.4&97.2&85.0 & 89.2		  \\	
	\textbf{TSB (ours)} (+ SWAD + MixStyle ) &  \underline{\textbf{90.8}}&  \underline{\textbf{84.5}}&97.1 &  \underline{\textbf{85.4}} &  \underline{\textbf{89.4}}   \\
		\bottomrule
	\end{tabular}
	\vspace{-4mm}
\end{table*}

\section{Ablation Studies on Style Balancing}
\vspace{-2mm}
We  first provide ablations studies on our style balancing (SB) module. In Step 2 of our style balancing procedure,   we proposed to move the style of the sample that has very similar statistics with other samples.  To validate the effectiveness of  this  idea, here we provide results with random sample selection; for domain $n$ satisfying $|\tilde{S}_{n,k}|>Q_k$, we randomly select $|\tilde{S}_{n,k}|-Q_k$ samples   to shift their styles to other domains. Table \ref{tab:effect_SB_class} shows  the results in a  cross-domain class imbalance scenario. The setup is exactly the same as in the main manuscript. The results of both Table \ref{tab:effect_SB_class} confirm the effectiveness of our sample selection strategy in SB compared to the random sampling strategy. 
 
 
\begin{table*}[h]
	\small
	\caption{\small  Effect of the proposed sample selection method in style balancing (SB) in \textbf{cross-domain class imbalanced PACS}.  }
	\centering
	\label{tab:effect_SB_class}
	\begin{tabular}{l|  c  ccc| c }
		\toprule  
		Methods  &Art  & Cartoon & Photo & Sketch & Average \\
				\midrule		
		 SB (+ Random sampling + MixStyle)  &43.37 &	54.02 &	54.91	&45.74&	49.51   \\
	 SB (+ Proposed sampling + MixStyle)&44.49&	55.57&	56.28&	44.93&	\underline{\textbf{50.32}} \\
				\midrule
		TSB (+ Random sampling + MixStyle)  &   43.43	&54.32	&54.91	&44.60 &49.32 \\
	TSB (+ Proposed sampling + MixStyle)  & 44.50	& 55.84	& 56.28& 	46.68	& \underline{\textbf{50.83}}  \\
		\bottomrule
	\end{tabular}
\end{table*}

%
%
%
%
%
%
%

\section{Ablation Studies on Test-Time Style Shifting}\label{app:Ts}

In this section, we provide ablation studies on our test-time style shifting (TS) module. 

\textbf{Variants of test-time style shifting.} We investigate the performance of  other  possible variants of TS. We consider two additional strategies for TS: first, instead of only  shifting the style of the test samples that have large style gaps with the source domains (as in the main manuscript),  we consider a scheme that \textit{shifts the styles of all samples} to   the nearest source domain (TS variant 1). We also consider a scheme that shifts the style of the sample to  the nearest sample among randomly selected 100 samples, based on the condition in equation (8) of the main manuscript (TS variant 2). Table \ref{tab:TS_baseline} compares the results of variants of TS. It can be first seen that TS variant 2 has lower performance compared to others, indicating that shifting the style to the nearest sample is less effective compared to the scheme that shifts the style to the nearest center of the source domain.  Shifting all the samples (TS variant 1) can improve the performance on Cartoon or Sketch domains, but suffers from performance degradation on Art or Photo; this indicates that it is better to keep the sample's original style when the gap with the source domain is small. In general, TS variant 1 achieves similar or lower performance compared to our TS strategy. 



\begin{table*}[ht]
	\small
	\caption{\small Comparison with other test-time style shifting (TS) variants in  original PACS.  }
	\centering
	\label{tab:TS_baseline}
	\begin{tabular}{l|  c  ccc| c }
		\toprule  
		Methods  &Art  & Cartoon & Photo & Sketch & Average \\
				\midrule		
		TSB variant 1 (shift all samples) (+ MixStyle)  &  82.71 &	81.66 &	95.55 &	78.81	 &\underline{\textbf{84.68}}  \\
				 TSB variant 2 (shift to the nearest sample)    (+ MixStyle) &83.60	 &79.57 &	96.15 &	77.25 &	84.14    \\
		\textbf{TSB proposed}  (+ MixStyle) &   83.62&80.07&96.15&78.66& 84.63   \\
		\midrule
	
	  TSB variant 1 (shift all samples)  (+ DSU) & 79.75	&80.18	&94.80 	&79.49	&83.55    \\
			  TSB variant 2 (shift to the nearest sample)     (+ DSU) &80.58& 80.14 & 95.83 & 77.92 & 83.62   \\
		\textbf{TSB proposed}  (+ DSU) &  80.73 &	80.69 &	95.83 &	79.47 	&\underline{\textbf{84.18}}   \\
		\bottomrule
	\end{tabular}
\end{table*}

\textbf{Location of TS module.} In the main manuscript, our TS module is applied at the output of the $2^{\text{nd}}$ residual block of ResNet-18, when training with PACS dataset. In Table \ref{tab:TS_location}, we applied the proposed TS module at different residual blocks. It is observed that applying TS module after   the $1^{\text{st}}$ block or the $2^{\text{nd}}$ block or the $3^{\text{rd}}$ block improves the performance. However, operating our TS module after the $4^{\text{th}}$  residual block significantly degrades the performance, which is straightforward since data are clustered according to the classes (regardless of the domains) at later layers.  

\begin{table*}[h]
	\small
	\caption{\small Effect of location of test-time style shifting (TS) module in original PACS.  }
	\centering
	\label{tab:TS_location}
	\begin{tabular}{l|  c  ccc| c }
		\toprule  
		Methods  &Art  & Cartoon & Photo & Sketch & Average \\
				\midrule		
					  SB (+ MixStyle)  & 83.48 & 79.07&96.15&73.74& 83.11  \\
				\midrule		

		  TSB (+ MixStyle)  (output of $1^{\text{st}}$ residual block)   &83.50	&79.11&	96.15&	75.67&	83.61   \\
				  TSB (+ MixStyle)  (output of $2^{\text{nd}}$  residual block)&  83.62&80.07&96.15&78.66& 84.63    \\
		  TSB (+ MixStyle) (output of $3^{\text{rd}}$ residual block)& 83.66	&79.80	&96.09&77.85&	84.35 \\
			  TSB (+ MixStyle)  (output of $4^{\text{th}}$ residual block)&  18.51	& 25.60	&18.84&17.89&	20.21 \\

		\bottomrule
	\end{tabular}
		\vspace{-3mm}
\end{table*}

\section{Detailed comparison with test-time adaptation baselines.}
This section provides a detailed comparison between our TSB and existing test-time adaptation works to clarify the difference. Different from the prior works, the unique contribution of our test-time style shifting is the effectiveness of handling arbitrary domains in the style-space, using feature-level style statistics, where the advantages are also confirmed via experiments. In Table \ref{tab:detailcomptta}, we compare the accuracy and inference time of our scheme with the recent test-time adaptation works, T3A \cite{iwasawa2021test} and Tent \cite{wang2020tent} using PACS dataset. We reimplemented them in our experimental setup, where the ``Baseline'' indicates training with ResNet-18 backbone without style augmentation. 

\textbf{Comparison with Tent \cite{wang2020tent}}: Although only the BN layers are perturbed in Tent, it still requires both the forward propagation for computing the entropy and the backpropagation for updating the BN layers. On the other hand, in our scheme, only one forward propagation is required along with the simple AdaIN process in the style space (without any backpropagation), further reducing the inference time compared to Tent. It is also worth mentioning that Tent requires a batch of test samples to compute the entropy, while our scheme can be operated sample-by-sample, which is another advantage of our test-time style shifting compared to Tent.

\textbf{Comparison with T3A \cite{iwasawa2021test}}: T3A does not require additional parameter updates during testing, which results in small inference time. However, it can be seen that our scheme achieves better generalization, especially for the Sketch domain that has a large style gap with other domains. This is the advantage of our test-time style shifting that is able to handle arbitrary styles. Moreover, T3A requires multiple test samples to continuously update the support set during testing, while our scheme can directly make a reliable prediction given only a single test sample.

\begin{table*}[h]
	\small
	\caption{\small  Detailed comparison with test-time adaptation baselines in original PACS.  }
	\centering
	\label{tab:detailcomptta}
	\begin{tabular}{l|  c  ccc| c | c}
		\toprule  
		Methods  &Art  & Cartoon & Photo & Sketch & Average & Inference time \\
				\midrule		
		 Tent (Baseline)	\cite{wang2020tent} &77.78&	78.03&	94.07 &66.27 &	79.04  & 40.08 ms\\
		 T3A (Baseline) \cite{iwasawa2021test} &73.83 &	77.65&	95.81 &	69.58 & 79.22 & 28.99 ms \\
		 TSB (Baseline) &80.60 &77.58 &	96.35 &	74.37 &	\underline{\textbf{80.22}} & 32.43 ms \\
				\midrule
		
		 Tent (+Mixstyle)	\cite{wang2020tent} &81.20 &	80.12 &	94.43 &74.80 &	82.64 & 40.08 ms\\
		 T3A (+Mixstyle) \cite{iwasawa2021test} &83.20 &	80.38 &	96.17 &	72.19 & 83.17 & 28.99 ms \\
		 TSB (+Mixstyle) &83.62 &80.07 &96.15 &	78.66 &	\underline{\textbf{84.63}} & 32.43 ms \\
		\bottomrule
	\end{tabular}
\end{table*}

\section{Comparison with BoDA \cite{yang2022multi}.}

We also compare our method with  the recent work \cite{yang2022multi} focusing on a  multi-domain setup with imbalanced datasets, in Table \ref{table;;boda}. The results on original PACS and cross-domain class imbalanced PACS again confirm the advantages of the proposed SB module over the recent work, BoDA \cite{yang2022multi}. By strategically handling the imbalance issue in DG (e.g., missing classes in each domain in class-imbalanced scenario), our style balancing can perform better than this baseline on both original and imbalanced PACS dataset.

\begin{table}[h]
	\scriptsize
	\caption{\small  Comparison with BoDA \cite{yang2022multi}.  }\label{table;;boda}
			\vspace{1mm}
	\begin{subtable}[!t]{1\linewidth}
	\centering
	\label{tab:com_boda}
		\caption{Results on original PACS.}
		\vspace{-1mm}
	\begin{tabular}{l|  c  ccc| c }
		\toprule  
		Methods  &Art  & Cartoon & Photo & Sketch & Average \\
				\midrule		
		{BoDA}  &74.46 &	76.54 &	95.27	& 67.72 &	78.50  \\
		SB (Baseline)   & 80.55&	77.16&	96.39&	71.68	&81.44 \\
		\textbf{SB}  (+ MixStyle) & 83.48 & 79.07&96.15&73.74&\underline{\textbf{83.11}} \\
		\bottomrule
	\end{tabular}
	\hfill
	\end{subtable}
		\begin{subtable}[!t]{1\linewidth}
	\centering
	\label{tab:com_boda}  
	\vspace{1.5mm}
	\caption{Results on cross-domain class imbalanced PACS.}
		\vspace{-1mm}
	\begin{tabular}{l|  c  ccc| c }
		\toprule  
		Methods  &Art  & Cartoon & Photo & Sketch & Average \\
				\midrule		
		{BoDA}  &53.56 &	63.27 &	94.97	& 60.92 &	68.18   \\
		SB (Baseline)   &65.04&	64.63 &	95.63	& 67.85 &	73.29   \\
		\textbf{SB}  (+ MixStyle) & 66.26&	64.46&	94.97&	71.44&	\underline{\textbf{74.28}} \\
		\bottomrule
	\end{tabular}
	\end{subtable}
					\vspace{-3mm}
\end{table}

\section{Results without Domain Labels}
Throughout the main manuscript, we described our algorithm using domain labels. In Table \ref{tab:wo_domain_label}, we show the performance of our scheme without any domain labels. Here, we provide pseudo domain labels using $k$-means clustering, where $k$ is set to be 3. We apply our SB and TS by utilizing the clustered domains with pseudo labels.  We  let $\alpha=2$ throughout all experiments in Table \ref{tab:wo_domain_label}. Experimental results show that both SB and TS are effective even without any domain labels.  The  performance of TS \textit{without} domain labels is sometimes even better compared to the case \textit{with} domain labels.  This indicates that  it is more  important to consider how the train samples are clustered in the style space,  rather than the original domain label, during  the TS process. 
\begin{table*}[h]
	\small
	\caption{\small  Performance without domain label on original PACS.}
	\centering
	\label{tab:wo_domain_label}
	\begin{tabular}{l|  c | cccc | c}
		\toprule  
		Methods & Reference &Art  & Cartoon & Photo & Sketch & Average \\
		\midrule	
		MixStyle  \cite{zhou2021domain}& ICLR'21&82.65&78.84	&96.09&	72.23	&82.45    \\
		    SB (+ MixStyle)  &   Ours&  83.72 &	79.34&	96.43	&73.22 &83.18  \\
	  TS  (+ MixStyle)  & Ours   & 83.10 &	80.99	&96.15&	78.11 	&84.59 \\
		\textbf{TSB} (+ MixStyle) &Ours&83.61 &	81.79 &	96.31	 &79.03  &\underline{\textbf{85.19}}  \\
				\midrule
		DSU  \cite{li2022uncertainty}& ICLR'22  &  81.78 & 78.66 & 95.91 & 76.75 & 83.27  \\
	 SB (+ DSU)&  Ours&81.92&	79.14&	95.95	&78.54	&83.89\\
	TS   (+ DSU)&  Ours&   80.16&	79.37	&94.91&	78.97 &	83.35 \\
		\textbf{TSB}  (+ DSU) & Ours &81.59&	80.01&	95.19&	79.16	&\underline{\textbf{83.99}}  \\
								\midrule
		EFDMix  \cite{zhang2022exact}  &   CVPR'22 &  83.35&	79.91	&96.67	&74.52&83.61  \\
 SB (+ EFDMix)&  Ours  &83.38 &	80.22	&96.81&	75.13&83.89  \\
 TS (+ EFDMix) &  Ours & 83.43 &	81.25	&96.26	&78.92&	84.96  \\
		\textbf{TSB}  (+ EFDMix)& Ours &   83.80 & 	81.57& 	96.49	& 79.05	& \underline{\textbf{85.23}} \\
		\bottomrule
	\end{tabular}
	\vspace{-3mm}
\end{table*}

\section{Effect of $\alpha$}\label{app:alpha}
 Recall that $\alpha$ is a hyperparameter that appears in equation (8) of the main manuscript.  In the main manuscript, we set $\alpha=3$ for all experiments for PACS and VLCS. However, this value may not be the optimal value for each domain/setup. In Table \ref{tab:varying_alpha},  we provide results on various  $\alpha$ values. When $\alpha$ is large  ($\alpha=5$), most of the test samples do not shift their styles; this reduces to the scheme with only SB. When $\alpha=0$, all the test samples move their styles to the nearest source domain, which can degrade the performance of specific domains (Art and Photo) but improves the performance of Cartoon and Sketch. One can also select the  $\alpha$ value by considering the extended validation set at feature-level; one can additionally generate new styles that have large style gaps with the current source domains, so that the extended set contains both samples that have small/large style gaps with the source domains. Nevertheless, whatever $\alpha$ we choose, we have additional performance improvement (or at least the same performance) compared to the case with no TS, confirming the advantage of our TS module.


\begin{table*}[ht]
	\small
	\caption{\small  Performance with varying $\alpha$ on original PACS: \textit{whatever $\alpha$ we choose, an additional performance gain can be obtained compared to no TS}.}
	\centering
	\label{tab:varying_alpha}
	\begin{tabular}{l| c | cccc | c}
		\toprule  
		Methods  & Reference & Art  & Cartoon & Photo & Sketch & Average \\
		\midrule	
		MixStyle  \cite{zhou2021domain}&  ICLR'21 & 82.54 & 79.42 & 95.88 & 74.06 & 82.98 \\
		SB (+ MixStyle)&  Ours & 83.48 & 79.07&\underline{\textbf{96.15}}&73.74&83.11  \\
	TSB (+ MixStyle) $(\alpha=0)$&  Ours  &  82.71 &	81.66 &	95.55 &	\underline{\textbf{78.81}}	 & 84.68    \\
	 		TSB (+ MixStyle)$(\alpha=2)$&  Ours  & 83.31 	& \underline{\textbf{81.81}}& 	96.01 & 	\underline{\textbf{78.81}}	& \underline{\textbf{84.99}}    \\
		 TSB (+ MixStyle) $(\alpha=3)$&  Ours  &\underline{\textbf{83.62}}&80.07&\underline{\textbf{96.15}}&78.66& 84.63    \\
		 	 	TSB (+ MixStyle) $(\alpha=4)$ &  Ours &    83.48&	79.10&	\underline{\textbf{96.15}}	&73.81 	&83.13  \\
TSB (+ MixStyle) $(\alpha=5)$ &  Ours &   83.48&  	79.07 &  	\underline{\textbf{96.15}}	&  73.74	&  83.11 \\	
		\bottomrule
	\end{tabular}
\end{table*} 

\section{Experiments on Office-Home Dataset}
In addition to the results on PACS, VLCS, Market1501 and GRID in the main manuscript, in Table \ref{tab:office_home}, we provide additional results on  Office-Home dataset \cite{venkateswara2017deep} with 4 domains and 65 classes.  We can observe a  performance gain via SB even in the original Office-Home dataset. The performance gain of TS is marginal since the style gaps between domains  are relatively small in Office-Home. Nevertheless, existing schemes can still benefit from the proposed SB and TS modules.

\begin{table*}[ht]
	\small
	\caption{\small  Performance   on original Office-Home dataset.}
	\centering
	\label{tab:office_home}
	\begin{tabular}{l| c | cccc | c}
		\toprule  
		Methods  & Reference & Art & Clipart & Product& Real world& Average \\
		\midrule	
		MixStyle  \cite{zhou2021domain}&  ICLR'21 &  57.99 & 53.04 & 73.64 & 74.98 & 64.91 \\
		 SB (+ MixStyle)&  Ours &  58.29 & 53.20 & 74.01 & 75.29 & 65.20 \\
	 \textbf{TSB} (+ MixStyle)&  Ours  &  58.27 & 53.41 & 74.05 & 75.33 & \underline{\textbf{65.27}}    \\
		\bottomrule
	\end{tabular}
\end{table*}

\section{Experiments on DomainNet Dataset}
We performed additional experiments on DomainNet \cite{peng2019moment} with 6 domains and 345 classes using ResNet-50. In Table \ref{tab:domainnet}, we compare our scheme with BoDA \cite{yang2022multi} on DomainNet dataset. The \textit{Baseline} indicates training with ResNet-50 backbone without style augmentation. The results show that our proposed TSB consistently outperforms Baseline and the recent work, BoDA, demonstrating the advantage of the proposed algorithm in a larger dataset.

\begin{table*}[ht]
	\small
	\caption{\small  Performance   on original DomainNet dataset.}
	\centering
	\label{tab:domainnet}
	\begin{tabular}{l | cccccc | c}
		\toprule  
		Methods   & Clipart & Infograph & Painting & Quickdraw & Real & Sketch & Average \\
		\midrule	
		Baseline & 60.77 & 25.63 & 50.31 & 12.46 & 62.33 & 48.80 & 43.38 \\
		BoDA (Baseline) \cite{yang2022multi} & 61.39 & 25.73& 50.01 & 12.50 & 62.20 & 48.72 & 43.43 \\
	 \textbf{TSB} (Baseline)  &  61.40 & 25.70 & 51.46 & 12.52 & 62.14 & 50.22 & \underline{\textbf{43.91}}    \\
		\bottomrule
	\end{tabular}
\end{table*}

 \section{Additional Experiments using ResNet-50}
In Table \ref{tab:ResNet50}, we show the results using ResNet-50. Other setups are exactly the same as in the main manuscript with ResNet-18. The results are consistent with all previous results, confirming  the  strong advantages of our SB and TS modules.

\begin{table*}[ht]
	\small
	\caption{\small  Performance comparison using ResNet-50 on original PACS.}
	\centering
	\label{tab:ResNet50}
	\begin{tabular}{l|  c | cccc | c}
		\toprule  
		Methods & Reference &Art  & Cartoon & Photo & Sketch & Average \\
		\midrule	
		MixStyle  \cite{zhou2021domain}& ICLR'21& 89.42&	81.94&	97.82&	76.04&	86.31  \\
	 SB (+ MixStyle) &   Ours&  89.782&	81.71&	97.80&	75.93&	86.31  \\
		\textbf{TSB} (+ MixStyle) &Ours& 89.92&	81.77&	97.80&	80.20&	 \underline{\textbf{87.42}}  \\
				\midrule
		DSU  \cite{li2022uncertainty}& ICLR'22  & 88.52	&82.32	&97.17&	76.42&	86.11   \\
		 SB (+ DSU)&  Ours&88.05	&82.90 	&97.62&	80.08	&87.16  \\
		\textbf{TSB}   (+ DSU)& Ours &88.11&82.94 	&97.59 &	82.04 &	 \underline{\textbf{87.67}}  \\
								\midrule
		EFDMix  \cite{zhang2022exact}  &   CVPR'22 &  89.68&	82.10	&97.84&	78.37 &	87.00 \\
	 SB (+ EFDMix) &  Ours  &  90.08&	81.75	&97.72&	78.17 &	86.93 \\
		\textbf{TSB} (+ EFDMix)  & Ours & 90.1 4&	81.80 	&97.66&	81.16 	& \underline{\textbf{87.69}}   \\
		\bottomrule
	\end{tabular}
\end{table*}

\section{Additional Experiments  for Instance Retrieval}
In this section, we provide the full version of Table 4 in the main manuscript. Table \ref{tab:retrieval_full} shows the corresponding result, confirming the effectiveness of the proposed style balancing and test-time style shifting strategies for instance retrieval, especially when they are used together. 

\begin{table*}[ht] 
	\small
	\caption{\small   Performance on  person re-ID task, using  Market1501 and GRID datasets. }
	\centering
	\label{tab:retrieval_full}
	\begin{tabular}{l|  c | cccc|c c c  c}
		\toprule  
		\multirow{2}{*}{Methods} & \multirow{2}{*}{Reference} &  \multicolumn{4}{c|}{Market $\rightarrow$  GRID} &  \multicolumn{4}{c}{GRID $\rightarrow$  Market} \\
		 &  & mAP & R1 & R5 &R10& mAP & R1 & R5 &R10\\
		\midrule
		MixStyle  & ICLR'21& 35.30	&26.67	&\textbf{44.53}	&53.07 & 5.25	&16.40	&30.05&	37.05\\
		 SB (+ MixStyle)&   Ours&  35.73 & 27.73 & 42.93& 52.00 &\textbf{5.70} & 17.70 & \textbf{31.90}&\textbf{39.65}\\ 
		 TS (+ MixStyle)& Ours   & 34.83 & 25.60 & 43.73& 50.67& 5.25 & 16.40 & 30.05 & 37.10  \\
		\textbf{TSB} (+ MixStyle)& Ours& \textbf{36.30}&	\textbf{28.27}&	42.93&	\textbf{55.47} &\textbf{5.70}	&\textbf{17.75}&	\textbf{31.90}&	\textbf{39.65}\\
		\midrule
		DSU & ICLR'22 &38.57	&30.40	&46.40	&53.07 & 4.45	&14.90&	27.65&	34.60 \\
			  SB (+ DSU)&  Ours& \textbf{41.47}&\textbf{33.33}&\textbf{48.80}&54.93&\textbf{5.25}&\textbf{16.75}&\textbf{31.65}&\textbf{38.85} \\
	 TS  (+ DSU)&  Ours&  37.27 & 28.00&46.13 & 55.73&4.40&14.75&27.35&34.60   \\
		\textbf{TSB}  (+ DSU)& Ours & 40.10& 30.67&	 48.00 	&\textbf{58.13} & \textbf{5.25}	&16.70	& 31.60 &	\textbf{38.85}\\

 
		\bottomrule
	\end{tabular}
\end{table*}

\section{Additional Experiments  in  Long-Tailed Imbalance Setting}\label{app:long}

We have performed additional experiments on the long-tailed imbalance setting   where the results are provided in Table \ref{tab:original_pacs__longtail}. The imbalance ratio, which represents the ratio between sample sizes of the most frequent and least frequent class, is set to 64. The results are consistent with the ones in the main manuscript. 

\begin{table}[ht]
	\small
	\caption{\small  Experimental results on long-tailed imbalance setting.  }
	\centering
	\label{tab:original_pacs__longtail}
	\begin{tabular}{l|  c | cccc | c}
		\toprule  
		Methods & Reference &Art  & Cartoon & Photo & Sketch & Average \\
		\midrule
		MixStyle  & ICLR'21& 73.49 & 76.75 & 86.17 &62.73 & 74.79    \\
	 SB (+ MixStyle)&   Ours& 76.46 & 75.30 & 88.20 & 61.81 & 75.44   \\
		  TS  (+ MixStyle)& Ours   & 73.68 & 76.75&86.17&69.04&77.53  \\
		\textbf{TSB}  (+ MixStyle)&Ours& 77.25& 75.64&88.20&69.04&\underline{\textbf{77.53}}\\
				\midrule
		DSU  & ICLR'22 & 75.47 &76.01 & 89.31 & 60.81 & 75.40  \\
	  SB (+ DSU)&  Ours&  73.66 & 76.17 & 90.93 & 67.51 & 77.07\\
	  TS (+ DSU) &  Ours&  74.54 & 76.43 & 89.16 & 66.55 & 76.67  \\
		\textbf{TSB}  (+ DSU)& Ours & 73.27 & 76.09 & 90.78& 68.92 &\underline{\textbf{77.26}}\\
		\bottomrule
	\end{tabular}
\end{table}

The results are consistent with the ones in our original manuscript, confirming the effectiveness of our algorithm in various imbalance scenarios including   the setup in  \cite{cao2019learning}. These results are also provided in Table 2 of Appendix.

\section{Algorithm in Pseudo Code}\label{app:pseudo}
Algorithm \ref{algo:SB} shows the sample selection process in style balancing. The process for test-time style shifting is provided in Algorithm \ref{algo:TS}.

	\begin{algorithm*}[ht]
	\small
	\caption{Sample Selection Process in Style Balancing (SB)}\label{algo:SB}
	\textbf{Input: } $\tilde{S}_{n,k}$ (samples in domain $n$ with class $k$, in a mini-batch) satisfying $|\tilde{S}_{n,k}|  > Q_k$ \\
	\textbf{Output: } $Z_{n,k}$, which contains $|\tilde{S}_{n,k}| -Q_k$ samples (with class $k$) to be shifted from domain $n$ to other source domains
		\begin{algorithmic}[1]
		\STATE $Z_{n,k} =\emptyset$, $E=0$
		\WHILE{ $E<|\tilde{S}_{n,k}| -Q_k$}
		    \FORALL{$s_{i}, s_{j}\in\tilde{S}_{n,k}$ $(i\neq j)$}	
		    \STATE Compute $d_{i,j} = \| \Phi(f(s_{i})) - \Phi(f(s_{j})) \|$
		    		    \ENDFOR
		    \STATE Choose two samples $(i^*,j^*)=\text{argmin}_{(i,j)}d_{i,j}$. \ \ \ \ 
		    \IF{$\text{min}\{d_{z, i^*}\}_{z=1, z\neq j^*}^{|\tilde{S}_{n,k}|} < \text{min}\{d_{z, j^*}\}_{z=1, z\neq i^*}^{|\tilde{S}_{n,k}|}$}
		    \STATE $Z_{n,k}\leftarrow Z_{n,k}\cup\{s_{i^*}\}$
		    \ELSE
		    \STATE $Z_{n,k}\leftarrow Z_{n,k}\cup\{s_{j^*}\}$
		    \ENDIF
		    \STATE $E\leftarrow E+1$
		    \ENDWHILE
	\end{algorithmic}
\end{algorithm*}

	\begin{algorithm*}[ht]
	\small
	\caption{Test-Time Style Shifting (TS)}\label{algo:TS}
	\textbf{Input: }Test sample $t$ and the corresponding feature $f(t)$ at a specific layer (where TS module is operated), $\Phi_{S}$ and $\Phi_{S_n}$ for all source domains $n\in\{1,2,\dots,N\}$, and $\alpha$.\\
	\textbf{Output: } New feature $\text{TS}(f(t))$. 
		\begin{algorithmic}[1]
		 \FOR{each test sample $t$}
		\STATE	Compute $\frac{1}{N}\sum_{n=1}^N \|\Phi(f(t)) - \Phi_{S_n} \|$
 \IF{$\frac{1}{N}\sum_{n=1}^N \|\Phi(f(t)) - \Phi_{S_n} \|      >  \alpha \Big( \frac{1}{N}\sum_{n=1}^N \|\Phi_S - \Phi_{S_n} \| \Big)$}
    \STATE $\Phi(f(t))_{\text{new}} = \Phi_{S_{n'}}$, where  $n'=\text{argmin}_n\|\Phi(f(t)) - \Phi_{S_n}\|$ \ // style shift to the nearest source domain
  \ELSE
    \STATE  $\Phi(f(t))_{\text{new}} = \Phi(f(t))$ \ \ \ \  // keep the original style
  \ENDIF
  	\STATE From  $\Phi(f(t))_{\text{new}} = [\mu(f(t))_{\text{new}},\sigma(f(t))_{\text{new}}]$,  
	\STATE compute $\text{TS}(f(t)) = \sigma(f(t))_{\text{new}}\frac{f(t)-\mu(f(t))}{\sigma(f(t))} + \mu(f(t))_{\text{new}}$ 
				\ENDFOR
	\end{algorithmic}
	\end{algorithm*}

\section{Other Implementation Details}\label{subsec;implement}
Our work is built upon the official setup of EFDMix  \cite{zhang2022exact}. Different from the original setting of EFDMix, for image classification tasks, we trained the model for 150 epochs with a mini-batch size of 128. We also randomly sampled the data from all source domains in each mini-batch. Other setups are exactly the same as in MixStyle  \cite{zhou2021domain}, DSU  \cite{li2022uncertainty} and EFDMix \cite{zhang2022exact}  when implementing each module; each module is activated with probability 0.5. Following the original setups, Mixstyle and EFDM are inserted after the 1st, 2nd and 3rd residual blocks  for PACS. For other datasets, Mixstyle and EFDM are inserted after the 1st and 2nd residual blocks.    DSU is inserted after 1st convolutional layer, max pooling, 1,2,3,4-th residual blocks.  Here, our SB module is operated  at the moment where MixStyle, DSU, EFDMix are first activated. The TS module is operated at first residual blocks during testing for VLCS, Office-Home and person re-ID task. We set $\alpha=3$ for all experiments on image classification tasks, while $\alpha=5$ is utilized for person re-ID task.


\end{document}